\documentclass[11pt]{article}

\usepackage[]{acl}

\usepackage{times}
\usepackage{latexsym}
\usepackage{graphicx}
\usepackage{amsmath}
\usepackage{multirow}
\usepackage{amsthm}
\usepackage{amsfonts}
\usepackage{bm}
\usepackage{booktabs}
\usepackage{algorithm}
\usepackage{algorithmic}
\usepackage{verbatim}

\usepackage[T1]{fontenc}

\usepackage[utf8]{inputenc}

\usepackage{microtype}

\title{HiSMatch: Historical Structure Matching based Temporal Knowledge Graph Reasoning}

\author{Zixuan Li\textsuperscript{1,2,3}\thanks{\ \ This work was done while the
first author was doing internship at Baidu Inc.}, Zhongni
Hou\textsuperscript{1,2}, Saiping Guan\textsuperscript{1,2}\thanks{\ \ Corresponding author.}, Xiaolong
Jin\textsuperscript{1,2}, Weihua Peng\textsuperscript{3},\\  \textbf{Long
Bai\textsuperscript{1,2}, Yajuan Lyu\textsuperscript{3}, Wei
Li\textsuperscript{3}, Jiafeng Guo\textsuperscript{1,2}, Xueqi
Cheng\textsuperscript{1,2}} \\
 \textsuperscript{1}School of Computer Science and Technology, University of Chinese Academy of Sciences; \\
 \textsuperscript{2}CAS Key Laboratory of Network Data Science and Technology, \\Institute of Computing Technology, Chinese Academy of Sciences;
 \textsuperscript{3}Baidu Inc. \\
 \texttt{\{lizixuan, houzhongni18z, guansaiping, jinxiaolong\}@ict.ac.cn}\\
 \texttt{\{pengweihua,lvyajuan\}@baidu.com}
 }

\begin{document}
\maketitle
\begin{abstract}
  A Temporal Knowledge Graph (TKG) is a sequence of KGs with respective
  timestamps, which adopts quadruples in the form of (\emph{subject},
  \emph{relation}, \emph{object}, \emph{timestamp}) to describe dynamic facts.
  TKG reasoning has facilitated many real-world applications via answering such
  queries as (\emph{query entity}, \emph{query relation}, \emph{?}, \emph{future
  timestamp}) about future. This is actually a matching task between a query and
  candidate entities based on their historical structures, which reflect
  behavioral trends of the entities at different timestamps. In addition, recent
  KGs provide background knowledge of all the entities, which is also helpful
  for the matching. Thus, in this paper, we propose the \textbf{Hi}storical
  \textbf{S}tructure \textbf{Match}ing (\textbf{HiSMatch}) model. It applies two
  structure encoders to capture the semantic information contained in the
  historical structures of the query and candidate entities. Besides, it adopts
  another encoder to integrate the background knowledge into the model. TKG
  reasoning experiments on six benchmark datasets demonstrate the significant
  improvement of the proposed HiSMatch model, with up to 5.6\% performance
  improvement in MRR, compared to the state-of-the-art baselines.
  \end{abstract}

\section{Introduction}
Knowledge Graphs (KGs), which store facts as triples in the form of
(\emph{subject}, \emph{relation}, \emph{object}), have been widely applied to
many NLP applications, such as question answering~\cite{lan2020query}, dialogue
generation~\cite{he2017learning} and recommendation~\cite{wang2019kgat}.
However, facts may constantly change over time. Temporal Knowledge Graphs (TKGs)
is a kind of KGs that describe such dynamic facts by extending each triple with
a timestamp as (\emph{subject}, \emph{relation}, \emph{object},
\emph{timestamp}). Usually, a TKG is represented as a sequence of KG snapshots.
The TKG reasoning task is to infer new facts from known ones, which primarily
has two settings, interpolation and extrapolation. The former attempts to
complete missing facts in history, while the latter aims to predict future facts
with historical facts. This paper focuses on the extrapolation setting, which is
more challenging and far from being solved~\cite{jin2020recurrent}. This task
can be seen as answering the query about the future facts (e.g.,
(\emph{COVID-19}, \emph{Infect}, \emph{?}, \emph{2022-8-1})) by selecting from
all the candidate entities. 

The key of answering the queries about future facts is to understand the history
thoroughly. All the existing models conduct reasoning based on substructures
extracted from the whole history. These substructures can be divided into two
types, i.e., query-related history~\cite{jin2019recurrent,zhu2021learning} and
candidate-related history~\cite{li2021temporal, li-etal-2022-complex,
han2021learning,deng2020dynamic}. The former contains the latest historical
facts related to the subject and relation in the query, which reflects the
behavioral trends of the subject concerning the query relation. The latter
contains all the latest historical facts of the candidates without considering
the query, which indicates the behavioral trends of all the entities. Both of
these two kinds of history are vital to TKG reasoning. Take the query
(\emph{COVID-19}, \emph{Infect}, ?, \emph{2022-8-1}) for example, the
query-related history contains facts like (\emph{COVID-19}, \emph{Infect}, *,
\emph{t}), where \emph{t} is before 2022-8-1. The candidate-related history of a
candidate \emph{A}, includes facts reflecting its own behaviors, like (\emph{A},
\emph{*}, \emph{*}, \emph{t}) or (\emph{*}, \emph{*}, \emph{A}, \emph{t}). In
the realistic situation, the occurrence of the fact (\emph{COVID-19},
\emph{Infect}, \emph{A}, \emph{2022-8-1}) is caused by the interactions between
these two kinds of history. However, existing models only focus on one kind of
history and underestimate the other, which limits their performance on TKG
reasoning. Overall, it still remains a challenge to model both two kinds of
history in a unified framework. 

To reduce the computational cost caused by the enormous facts in history, these
two kinds of history usually contain one hop facts of the centered entities.
Thus, they cannot model the high-order associations among the entities, which is
also vital to TKG reasoning. 

Motivated by these, we consider both query-related history and candidate-related
history under a matching framework and propose the \textbf{Hi}storical
\textbf{S}tructure \textbf{Match}ing (\textbf{HiSMatch}) model. Specifically, it
applies two structure encoders to model the semantic information in the above
two kinds of historical structures, respectively. Then, it obtains the matching
scores. Both of these two structure encoders contain three components: (1) a
structure semantic component to model the structure dependencies among
concurrent facts at the same timestamp; (2) a time semantic component to model
the time numerical information of the historical facts; (3) a sequential pattern
component to mine the behavioral trends from the temporal order information.
Additionally, to model the high-order associations among the entities, we
consider the most recent KGs as the background knowledge of each query and apply
a GCN-based background knowledge encoder to obtain more informative entity
representations for the two structure encoders.  

Our contributions are summarized as follows:
\begin{itemize}

\item We first advocate the importance of modeling both query-related and
candidate-related history for TKG reasoning and transform the task into a
matching problem between them.

\item To solve this problem, we propose HiSMatch to comprehensively capture the
information in both historical structures via modeling the structure
dependencies among concurrent facts, the time numerical information of
historical facts and the temporal order among facts. HiSMatch complementally
captures high-order associations among entities by modeling the recent
background knowledge.

\item Extensive experiments on six commonly used benchmarks demonstrate that
HiSMatch achieves significantly better performance (up to 5.6\% improvement
in MRR) on the TKG reasoning task.
\end{itemize}

\begin{figure*}[t]
  \centering
  \includegraphics[width=0.7\linewidth]{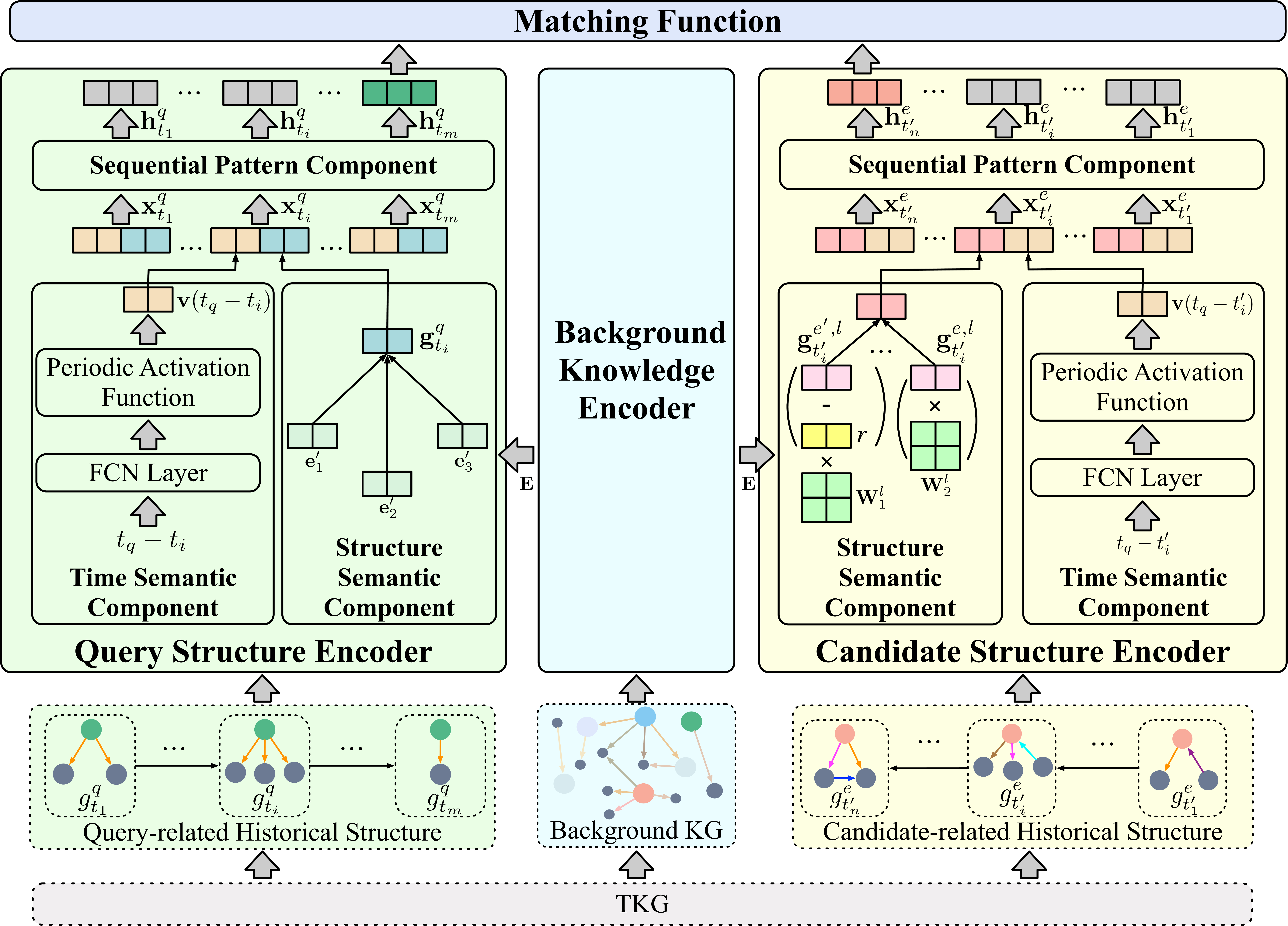}
  \caption{An illustrative diagram of the proposed HiSMatch model.}
  \label{model_fig}
  \vspace{-5mm}
\end{figure*}

\section{Related Work}
\textbf{TKG Reasoning under the interpolation setting} focuses on completing the
missing facts at past timestamps
\cite{liao2021learning,goel2020diachronic,wu2020temp,han2020dyernie,jiang2016encoding,dasgupta2018hyte,garcia2018learning,xu2021time}.
For example, TTransE \cite{leblay2018deriving} extends the idea of TransE
\cite{bordes2013translating} by adding the temporal order constraints among
facts. Also, HyTE \cite{dasgupta2018hyte} projects the entities and relations to
time-related hyperplanes to generate time-aware representations.
TNTComplEx~\cite{lacroix2020tensor} performs 4th-order tensor factorization to
get the time-aware representations of entities. However, they cannot obtain the
representations of the unseen timestamps and are not suitable for the
extrapolation setting.

\textbf{TKG Reasoning under the extrapolation setting} aims to predict facts at
future timestamps. According to the historical structure the models focus on,
the existing models can be categorized into two groups: query-based and
candidate-based models. 

\emph{Query-based models} focus on modeling the query-related history. For
example, RE-NET~\cite{jin2020recurrent} models the query-related subgraph
sequence. GHNN~\cite{han2020graph} introduces the temporal point process to
model the precise time information and takes the 1-hop subgraphs of the query
entity into consideration. CyGNet~\cite{zhu2021learning} captures repetitive
patterns by modeling repetitive facts with the same subject and relation to the
query. xERTE~\cite{han2020explainable} learns a dynamic pruning procedure to
find the query-related subgraphs. CluSTeR~\cite{li2021search} and TITer
~\cite{sun2021timetraveler} both adopt reinforcement learning to discover
query-related paths in history. 

\emph{Candidate-based models} encode the latest historical facts of all the
candidate entities without considering the query, and query are considered only
in the decoding phase. RE-GCN and its extension CEN~\cite{li2021temporal, li-etal-2022-complex} designs an evolutional model
to get the representations of all the candidates by modeling history at a few
latest timestamps. TANGO~\cite{han2021learning} utilizes neural ordinary
differential equations to model the structure information for each candidate
entity. Glean~\cite{deng2020dynamic} introduces unstructured textual information
to enrich the candidate-related history.

Above all, none of the existing models focus on both two kinds of history in a
unified framework. HiSMatch considers these two kinds of history under the
matching framework and takes the advantages of both kinds of models.


\section{Problem Formulation}
A TKG $G=\{G_0, ..., G_t, ..., G_T\}$ is a sequence of KGs, each of which
contains facts occurred at timestamp $t$, i.e., $G_t=\{\mathcal{E}, \mathcal{R},
\mathcal{F}_t\}$, where $\mathcal{E}$ is the set of entities, $\mathcal{R}$ is
the set of relations and $\mathcal{F}_{t}$ is the set of facts that occurred at
$t$. Each fact is a quadruple $(e_s,r,e_o,t)$, where $e_s,e_o \in \mathcal{E}$
and $r \in \mathcal{R}$. For each fact in TKG, we add the inverse quadruple
$(e_o,r^{-1},e_s,t)$ into TKG, correspondingly. The TKG reasoning task aims to
predict the missing object via answering a query $q = (e_q,r_q,?,t_q)$ with the
historical KGs given. Note that, when predicting the missing subject of a query
$q=\{?, r_q, e_q, t_q\}$, we can convert the query into $q=\{e_q, r_q^{-1}, ?,
t_q\}$.

\section{The HiSMatch model}
HiSMatch aims to captures the semantic similarity contained in the query-related
and candidate-related historical structures. For each query time, it first
embeds the background knowledge into the initial entity representations. With
the initial representations as input, it maps the semantic information in these
two historical structures into the vectorized representations of structures.
Based on the structure representations, matching scores are calculated.

Thus, as shown in Figure~\ref{model_fig}, HiSMatch consists of four parts: the
query structure encoder, the candidate structure encoder, the background
knowledge encoder, and the matching function. First, two kinds of historical
structures and a background knowledge graph are derived from the TKG. Then, the
background knowledge encoder gets the representations of the entities with the
background knowledge graph as input
(Section~\ref{background_knowledge_encoder}). With the learned representations
as input, two structure encoders use three components to integrate three kinds
of semantic information into the representations of query-related structure and
candidate-related structure, respectively (Section
~\ref{query_structure_encoder} and ~\ref{candidate_structure_encoder}). Finally,
the matching function calculates the scores between the query and candidates
based on the representations of their historical structures
(Section~\ref{matching_function}).


\subsection{Query Structure Encoder}\label{query_structure_encoder}
The query-related historical structure should reflect the behavioral trends of
the query. Motivated by this, for a query $q=(e_q, r_q, ?, t_q)$, the
query-related historical structure consists of the latest historical facts with
the same subject $e_q$ and relation $r_q$. These facts co-occuring at the same
timestamp $t$ form a subgraph $g^{q}_{t}$ centered on $e_q$. Then, we obtain a
subgraph sequence $\{g^{q}_{t_1}, ..., g^{q}_{t_i}, ..., g^{q}_{t_m}\}$, where
$t_1<...<t_i<...<t_m<t_q$ and $m$ is the maximum length of the sequence.

Three kinds of information are vital in the above historical structure, namely,
the structure semantic information of each subgraph, the time numerical
information of each subgraph, and the temporal order information across
subgraphs. To model these three kinds of information, we design three components
as follows:

\textbf{Structure Semantic Component.} 
The structure semantic information captures the associations among the query
entity and other entities through the query relation and implies possible answer
entities. Since all the concurrent facts, which having the same subject and
relation to the query, form a one-hop homogeneous graph, we simply perform mean
pooling over all the neighbor entities in each subgraph $g^q_{t_i}$ to get the
structure semantic representation $\mathbf{g}_{t_i}^{q}$ of the subgraph,

\begin{equation}
  \mathbf{g}_{t_i}^{q}=\frac{1}{|N_{t_i}^{q}|}\sum_{e\in N_{t_i}^{q}}\mathbf{e},
  \end{equation}
where $N_{t_i}^q$ is the set of the neighborhood of the query entity $e_q$ in
$g^q_{t_i}$ and $\mathbf{e}$ is the representation of each entity $e$ calculated
by the background knowledge encoder (see Section~\ref{background_knowledge_encoder}).  

\textbf{Time Semantic Component.}\label{tscomponent} Previous
works~\cite{jin2019recurrent,jin2020recurrent} only consider the temporal order
of the facts but ignore their time numerical information. A much earlier fact
and a recent one contribute equally when they have the same order in the
subgraph sequence. Actually, the recent fact is more important. Motivated by
this, we model the time numerical information by encoding the time interval $d =
t_q-t_i$, into the time representation $\mathbf{v}(d)$. However, giving each
time interval a learnable time representation always meets the time sparsity
problem (i.e., the time interval used in the test phase may not exist in the
training phase). Thus, we model any time interval by rescaling a learnable time
unit $\mathbf{w}_t$ with a time bias $\mathbf{b}_t$, 
\begin{equation}
  \mathbf{v}(d)=cos(d\mathbf{w}_t+\mathbf{b}_t).
\end{equation}

Since some facts occur periodically, such as elections, we additionally apply
the periodic activation function, i.e., cosine function, on $\mathbf{v}(d)$.

\textbf{Sequential Pattern Component.} Furthermore, the temporal order
information in the subgraph sequence implies sequential patterns of the query
entity. To integrate the sequential patterns into the representation of the
query, we use Gated Recurrent Unit (GRU) to model the subgraph sequence. First,
for every timestamp $t_i$ ($i={1,2,...,m}$), we concatenate structure semantic
representation and the time semantic representation from the above two
components as the input of GRU, 
\begin{equation}
  \mathbf{x}_{t_i}^q=[\mathbf{g}_{t_i}^q; \mathbf{v}(d)].
\end{equation}

Then these representations $\{\mathbf{x}_{t_1}^q, ..., \mathbf{x}_{t_i}^q, ...,
\mathbf{x}_{t_m}^q \}$ are fed into GRU recursively, 
\begin{equation}
\mathbf{h}_{t_{i}}^q=GRU(\mathbf{h}_{t_{i-1}}^q, \mathbf{x}_{t_{i}}^{q}),
\end{equation}
where $i\in \{1, 2, ..., m\}$ and $\mathbf{h}_{t_0}^q$ is the randomly
initialized hidden representations for GRU. The final representation of query
$(e_q, r_q, ? , t_q)$ is the ouput of GRU at the final step, i.e.,
$\mathbf{h}_{t_{q}}^q = \mathbf{h}_{t_{m}}^q$.

\subsection{Candidate Structure Encoder} \label{candidate_structure_encoder} The
candidate-related historical structure reflects the behavioral trends of each
candidate entity. For each candidate entity $e$, we use its 1-hop subgraphs at
latest historical timestamps to form a subgraph sequence $\{g_{t'_1}^{e},...,
g_{t'_i}^{e},..., g_{t'_n}^{e}\}$. $n$ is the maximum length of the sequence.
Actually, this structure is similar to the query-related historical structure,
the difference is that each subgraph in the sequence is multi-relational.
Therefore, we use an encoder similar to the query structure encoder, except the
calculation of the structure semantic representation of each subgraph. More
specifically, we adopt the CompGCN~\cite{vashishth2019composition} instead of
the mean pooling operation, to capture the semantic information of different
relations\footnote{Note that, the CompGCN layers can be replaced by other
relation-aware GCNs. We further analyze them in Section\ref{sec: gcn}}. The
representation of the candidate entity $e$ at timestamp $t_i$, is calculated by
a CompGCN with $\omega_1$ layers. Thus, the representation of the $(l+1)$-th
layer is

\begin{equation}
  \begin{split}
\mathbf{h}_{t'_i}^{e, l+1}=&f(\frac{1}{c_e}\sum_{e'\in N_{t'_i}^{e}}\mathbf{W}_1^l(\mathbf{h}_{t'_i}^{e', l}-\mathbf{r})\\
&+\mathbf{W}_2^l \mathbf{h}_{t'_i}^{e, l}),
  \end{split}
  \label{eq:compgcn}
 \end{equation}
where $\mathbf{r}$ is the representation of relation $r$;
$\mathbf{h}_{t'_i}^{e', l}$ denotes the $l$-th layer representation of entity
$e'$ at $t'_i$ timestamp; $\mathbf{W}^l_1$, $\mathbf{W}^l_2$ are the weight
matrices of the $l$-th layer; $c_e$ is a normalization constant, which equals to
the in-degree of entity $e$. Note that the input representations of all entities
are also calculated by the background knowledge encoder, which will be
introduced in Section~\ref{background_knowledge_encoder}.  

Then, the structure semantic representation of the subgraph $g_{t'_i}^{e}$,
i.e., $\mathbf{g}_{t'_i}^{e}$, equals to the representation of the centered
candidate $e$ from the last layer of CompGCN, i.e., $\mathbf{g}_{t'_i}^{e}=
\mathbf{h}_{t'_i}^{e,\omega_1}$.
  

Similar to the query structure encoder, we use another GRU to model the subgraph
sequence. The input of GRU at timestamp $t'_i$ is 
\begin{equation}
  \mathbf{x}_{t'_i}^{e}= [\mathbf{g}_{t'_i}^{e}; \mathbf{v}(d)],
\end{equation}
where $d=t_q-t'_i$ and $\mathbf{v}(d)$ is the time interval representation calculated by a shared
time semantic component introduced in Section~\ref{tscomponent}. Finally, the
output of GRU at the last step $t'_{n}$ is used as the representation of
candidate entity $e$ at $t_q$, i.e., $\mathbf{h}_{t_q}^{e}=\mathbf{h}_{t'_{n}}^{e}$.

\begin{table*}
  \scriptsize 
  \centering
  \begin{tabular}{lrrrrrrrr}
  \toprule
  Datasets  & $|\mathcal{V}|$  & $|\mathcal{R}|$  & $|\mathcal{E}_{train}|$ &
  $|\mathcal{E}_{valid}|$ &$|\mathcal{E}_{test}|$   & Time granularity &Snapshot numbers\\
  \midrule
  ICEWS14       &7,128    &230     &74,845    &8,514   &7,371    &24 hours    &365\\
  ICEWS14*       &7,128   &230     &63,685    &13,823   &13,222    &24 hours   &365\\
  ICEWS18       &23,033   &256     &373,018   &45,995  &49,545  &24 hours        &365\\
  ICEWS05-15    &10,094   &251     &368,868   &46,302  &46,159  &24 hours       &4017\\
  GDELT         &7,691    &240     &1,734,399  &238,765 &305,241 &15 mins       &2975\\
  WIKI          &12,554   &24      &539,286   &67,538  &63,110     &1 year      &232\\
  \bottomrule
  \end{tabular}
  \caption{Statistics of the datasets ($|\mathcal{E}_{train}|$,
  $|\mathcal{E}_{valid}|$, $|\mathcal{E}_{test}|$ are the sizes of training,
  validation, and test sets.).}
  \label{table:datasets}
  \end{table*}

  \subsection{Background Knowledge Encoder}\label{background_knowledge_encoder}
  The above two historical structures are local information centered
  on the query entity or the candidate entity, which focus on describing the
  behavioral trends of the entities. However, these two kinds of structures may
  miss some important entities that have high-order associations with the query
  entity or the candidate entity in the whole TKG. Since the recent history is
  more important, for query timestamp $t_q$, we gather the latest $k$ KGs into a
  cumulative graph $\mathcal{G}_{t_q}$, called background knowledge graph.
  Formally, $\mathcal{G}_{t_q}=\{\mathcal{E}, \mathcal{R},
  \hat{\mathcal{F}}_{t_q}=\{(e_s, r, e_o)| (e_s, r, e_o, t)\in \mathcal{F}_t,
  t_q-k \le t<t_q\}\}$, where $\hat{\mathcal{F}}_{t_q}$ is a set of facts. We
  adopt another CompGCN with $\omega_2$ layers to model it since it is also a
  multi-relational graph. The representations of all entities are calculated as
  follows, 
  \begin{equation}
    \mathbf{E}= CompGCN(\mathbf{E'}, \mathbf{R}, \mathcal{G}_{t_q}),
  \end{equation}
  where $\mathbf{E'}$ is the randomly initialized entity representation matrix
  and $\mathbf{E}$ is used as the input entity representation matrix of the
  aforementioned two structure semantic components. $\mathbf{R}$ is the relation
  representation matrix, which is shared with the structure semantic component.
  For the entities that have no facts in the background knowledge graph, an
  self-loop operation is conduct to get its representation. Note that, the
  background knowledge graph changes along the query time and $\mathbf{E}$ is
  different for different $t_q$.
  
  
  

  \subsection{Matching Function} \label{matching_function} With the
  representation $\mathbf{h}_{t_{q}}^q$ of the query and the representation
  $\mathbf{h}_{t_{q}}^e$ of each candidate entity $e$ at timestamp $t_q$ as
  input, the matching function calculates the score of the quadruple $(e_q, r_q,
  e, t_q)$. As previous work~\cite{vashishth2019composition, li2021temporal}
  shows the convolutional score functions get good performance on reasoning
  tasks, ConvTransE~\cite{shang2019end} is chosen as the matching fucntion,
  which contains 1D convolution and fully-connected layers, denoted by
  $ConvTransE(\cdot)$. To describe the behavioral information of the query
  entity in the query representation, an sum-up operation is performed between
  $\mathbf{h}_{t_q}^{q}$ and $\mathbf{h}_{t_q}^{e_q}$. Then, the score for eacht
  candidate entity $e$ is calculated as follows: 
  \begin{equation}
    \begin{split}
      \phi(e_q, &r_q, e, t_q)=\\
      &\sigma(\mathbf{h}_{t_q}^{e}ConvTransE(\mathbf{h}_{t_q}^{q}+\mathbf{h}_{t_q}^{e_q},\mathbf{r_q})),
    \end{split}
    \label{eq:scores}
   \end{equation}
  where $\sigma(\cdot)$ is the sigmoid function.
  
  \subsection{Training Details}
  The training objective is to minimize the cross-entropy loss:
  \begin{equation}
  L(\Theta)= \sum^{T}_{t=0} \sum_{(e_s,r,e_o,t)\in \mathcal{F}_{t}}\sum_{e\in\mathcal{E}}{y}_{t}^{e} \log \phi(e_s, r, e, t),
  \end{equation}
  where $T$ is the number of timestamps in the training set; ${y}_{t}^{e}=1$ if
  $e$ equals to $e_o$, otherwise $0$; $\phi(e_s, r, e, t)$ is the matching score
  between the query $(e_s, r, ?, t)$ and the candidate entity $e$; $\Theta$ are
  all the model parameters.

\section{Experiments}
We compare HiSMatch with a number of baselines on six datasets to validate its
effectiveness. In addition, we conduct ablation study to analyze the importance
of its different parts. We also evaluated the effects of different kinds of GCN
layers in the candidate structure encoder and the background knowledge encoder.
Besides, we study the maximum time interval that HiSMatch models.
\subsection{Experimental setup}
\subsubsection{\bf{Datasets}} To evaluate the effectiveness of HiSMatch, we use
the following six benchmark TKGs: ICEWS14~\cite{li2021temporal},
ICEWS14*~\cite{han2020explainable}, ICEWS18~\cite{jin2020recurrent},
ICEWS05-15~\cite{li2021temporal}, GDELT~\cite{jin2020recurrent} and
WIKI~\cite{leblay2018deriving}. The first four datasets with the time
granularity of 24 hours were extracted from the large-scale event-based
database, Integrated Crisis Early Warning System. The ICEWS14, ICEWS14* and
ICEWS18 datasets contain events in 2014 and 2018, respectively, and the
ICEWS05-15 dataset contains events occurred from 2005 to 2015. GDELT is
extracted from the global database of events, language, and
tone~\cite{leetaru2013gdelt}, which has a fine-grained time granularity of 15
minutes. WIKI is a TKG with the largest time granularity of one year. The
statistics of the datasets are listed in Table~\ref{table:datasets}.

\subsubsection{\bf Evaluation Metrics} 
We employ widely used $Hits@N$ and Mean Reciprocal Rank (MRR) to evaluate the
performance of the models. $Hits@N$ measures the proportion of correct entities
whose scores rank less than or equal to $N$. In this paper, $N \in \{1, 3,
10\}$, i.e., the results in terms of $Hits@1$, $Hits@3$, and $Hits@10$ are reported.
MRR measures the average of these reciprocal ranks and is the most typical
metric for TKG reasoning. Previous
work~\cite{han2020explainable,han2021learning,li2021search,li2021temporal}
points out that the traditional filtered setting is flawed as it ignores the
time of the fact and filters all facts with the same entity and relation before
ranking. Actually, only the facts occurring at the same time should be filtered.
Thus, we calculate the results under the more reasonable time-aware filtered
setting following ~\citet{sun2021timetraveler,han2021learning}, which only
filters out the quadruples occurring at the query time.

\subsubsection{\bf Baselines}
The HiSMatch model is compared with three kinds of models: KG reasoning models,
interpolation TKG reasoning models and extrapolation TKG reasoning models. For
the KG reasoning models, DistMult~\cite{yang2015embedding},
ComplEx~\cite{trouillon2016complex}, ConvE~\cite{dettmers2018convolutional},
ConvTransE~\cite{shang2019end}, RotatE~\cite{sun2018rotate} are compared. For
the TKG reasoning models, HiSMatch is compared to the interpolation TKG
reasoning models, including TTransE~\cite{leblay2018deriving},
TA-DistMult~\cite{garcia2018learning}, DE-SimplE~\cite{goel2020diachronic} and
TNTComplEx~\cite{lacroix2020tensor}. Besides, as for the extrapolation TKG
reasoning models, we choose the newest seven baselines including
TANGO-DistMult~\cite{han-etal-2021-learning-neural},
TANGO-Tucker~\cite{han-etal-2021-learning-neural},
xERTE~\cite{han2020explainable}, TITer~\cite{sun2021timetraveler},
CyGNet~\cite{zhu2021learning}, RE-NET~\cite{jin2020recurrent} and
RE-GCN~\cite{li2021temporal}.

\subsubsection{\bf Implementation Details} 
The dimensions of the entities and relations are set to 128, and the dimension
of the time semantic representation is set to 32 for all the datasets. For the
structure semantic encoders, the optimal lengths of historical structures of
query $m$ and candidate entities $n$ are equal in this paper. For ICEWS14,
ICEWS18, ICEWS05-15 and GDELT, they are set to 5; while 6 for ICEWS14* and 1 for
WIKI; the number of layers $\omega_1$ of the CompGCN is set to 1 for GDELT and 2
for the other datasets; the GRU layers is set to 1 for all the datasets and the
output dimension of the GRU unit is set to 128. For the background knowledge
encoder, the latest KG number $k$ is experimentally set to 4, 1, 2, 1, 2, 2 for
ICEWS14, ICEWS18, GDELT, ICEWS14*, ICEWS05-15, and WIKI, respectively; we set
the dropout rate for each layer to 0.2 and the layer of CompGCN in the
background knowledge encoder, $\omega_2$, to 2, for all the datasets. For the
matching function, the number of kernels is set to 50, the kernel size is set to
$2\times3$, and the dropout rate is set to $0.2$, for all the datasets.
Adam~\cite{kingma2014adam} is adopted for parameter learning with the learning
rate 0.001. All the experiments are carried out on 32G Tesla V100.


\begin{table*}[htb]
  \scriptsize 
\centering
\begin{tabular}{lrrrrrrrrrrrrrrrr}
\toprule
\multirow{2}{*}{Model} &\multicolumn{4}{c}{ICE14} &\multicolumn{4}{c}{ICE05-15} &\multicolumn{4}{c}{GDELT}\\
\cmidrule(r){2-5}  \cmidrule(r){6-9} \cmidrule(r){10-13}  &MRR &H@1 &H@3 &H@10 &MRR &H@1 &H@3 &H@10 &MRR &H@1 &H@3 &H@10\\
\midrule
DistMult\!\!\!&15.44&10.91&17.24&23.92&17.95&13.12&20.71&29.32&8.68&5.58&9.96&17.13          \\
ComplEx\!\!\!&32.54&23.43&36.13&50.73&32.63&24.01&37.50&52.81&16.96&11.25&19.52&32.35\\
ConvE\!\!\!&35.09&25.23&39.38&54.68&33.81&24.78&39.00&54.95&16.55&11.02&18.88&31.60        \\
ConvTransE\!\!\!&33.80&25.40&38.54&53.99&33.03&24.15&38.07&54.32&16.20&10.85&18.38&30.86  \\
RotatE\!\!\!&21.31&10.26&24.35&44.75&24.71&13.22&29.04&48.16&13.45&6.95&14.09&25.99  \\
\midrule
TTransE\!\!\! &13.72 &2.98 &17.70 &35.74  &15.57&4.80&19.24&38.29 &5.50&0.47&4.94&15.25    \\
TA-DistMult\!\!\!  &25.80 &16.94 &29.74 &42.99  &24.31&14.58&27.92&44.21&12.00&5.76&12.94&23.54  \\
DE-SimplE\!\!\!  &33.36 &24.85 &37.15 &49.82&35.02&25.91&38.99&52.75&19.70&12.22&21.39&33.70    \\
TNTComplEx\!\!\! &34.05&25.08&38.50&50.92  &27.54&9.52&30.80&42.86&19.53&12.41&20.75&33.42\\
TANGO-DistMult\!\!\!&-&-&-&-&40.71&31.23&45.33&58.95&-&-&-&- \\
TANGO-Tucker\!\!\!&-&-&-&-&  42.86&32.72&48.14&62.34&-&-&-&-          \\
xERTE\!\!\!&40.02&32.06&44.63&56.17 &46.62&37.84&52.31&63.92&18.09&12.30&20.06&30.34    \\
TITer\!\!\!&40.87&32.28&45.45&57.10&47.69&37.95&52.92&65.81&15.46&10.98&15.61& 24.31        \\
CyGNet &35.05&25.73&39.01&53.55&36.81&26.61&41.63&56.22&18.48&11.52&19.57&31.98 \\
RE-NET\!\!\!&36.93&26.83&39.51&54.78&43.32&33.43&47.77&63.06&19.62&12.42&21.00&34.01    \\
RE-GCN\!\!\!&40.39&30.66&44.96&59.21&48.03&37.33&53.85&68.27&19.64&12.42&20.90&33.69  \\
\midrule
HiSMatch \!\!\!    &\bf{46.42}&\bf{35.91}&\bf{51.63}&\bf{66.84} &\bf{52.85} &\bf{42.01} &\bf{59.05} &\bf{73.28} &\bf{22.01} & \bf{14.45} &\bf{23.80}  &\bf{36.61}\\  
\bottomrule
\end{tabular}
\caption{Performance (in percentage) on ICEWS14, ICEWS05-15 and GDELT. We
average the results of HiSMatch over five runs and the best results are in
bold.}
\label{table:result1}
\end{table*}

\begin{table*}[htb]
\scriptsize 
\centering
\begin{tabular}{lrrrrrrrrrrrr}
\toprule
\multirow{2}{*}{Model} &\multicolumn{4}{c}{ICE14*} &\multicolumn{4}{c}{ICE18}&\multicolumn{4}{c}{WIKI} \\
\cmidrule(r){2-5}  \cmidrule(r){6-9} \cmidrule(r){10-13} &MRR &H@1 &H@3 &H@10 &MRR &H@1 &H@3 &H@10&MRR &H@1 &H@3 &H@10 \\
\midrule
DistMult\!\!\! &16.16&11.42&17.94&25.30&11.51&7.03&12.87&20.86&10.89&8.92&10.97&16.82     \\
ComplEx\!\!\!&21.28&14.49&23.11&35.20&22.94&15.19&27.05&42.11&24.47&19.69&27.28&34.83       \\
ConvE\!\!\!&34.50&24.83&38.56&53.88&24.51&16.23&29.25&44.51&14.52&11.44&16.36&22.36       \\
ConvTransE\!\!\!&33.47&25.15&38.15&53.30&22.11&13.94&26.44&42.28&10.60&8.67&11.94&16.93  \\
RotatE\!\!\!&20.88&10.26&23.90&44.03&12.78&4.01&14.89&31.91&46.10&41.89&49.65&51.98  \\
\midrule
TTransE\!\!\!&13.43&3.11&17.32&34.55&8.31&1.92&8.56&21.89 &29.27&21.67&34.43&42.39 \\
TA-DistMult\!\!\!&26.47&17.09&30.22&45.41&16.75&8.61&18.41&33.59&44.53&39.92&48.73&51.71\\
DE-SimplE\!\!\!&32.67&24.43&35.69&49.11&19.30&11.53&21.86&34.80&45.43&42.60&47.71&49.55\\
TNTComplEx\!\!\!&32.12&23.35&36.03&49.13&21.23&13.28&24.02&36.91&45.03&40.04&49.31&52.03\\
TANGO-DistMult\!\!\!&24.70&16.36&27.26&41.35&26.65&17.92&30.08&44.09&51.15&49.66&52.16&53.35\\
TANGO-Tucker\!\!\! &26.25&17.30&29.07&44.18&28.68&19.35&32.17&47.04&50.43&48.52&51.47&53.58\\
xERTE\!\!\!&40.79&32.70&45.67&57.30&29.98&22.05&33.46&44.83&71.14&68.05&76.11&79.01\\
TITer\!\!\! &41.73&32.74&46.46&58.44&29.98&22.05&33.46&44.83&75.50&72.96&77.49&79.02\\
CyGNet&32.73&23.69&36.31&50.67&24.93&15.90&28.28&42.61&33.89&29.06&36.10&41.86\\
RE-NET\!\!\!&38.28&28.68&41.43&54.52&28.81&19.05&32.44&47.51&49.66&46.88&51.19&53.48\\
RE-GCN\!\!\!&41.78&31.58&46.65&61.51&30.58&21.01&34.34&48.75&77.55&73.75&80.38&83.68\\
\midrule
HiSMatch\!\!\!    &\bf{45.82}&\bf{35.84}&\bf{50.79}&\bf{65.08} &\bf{33.99} &\bf{23.91} &\bf{37.90} &\bf{53.94} &\bf{78.07} &\bf{73.89} &\bf{81.32} &\bf{84.65} \\ 
\bottomrule
\end{tabular}
\caption{Performance (in percentage) on ICEWS14*, ICESW18 and WIKI.}
\label{table:result2}
\end{table*}

\subsection{Experimental Results} \label{exp_result} The experimental results of
HiSMatch and all the baselines on TKG reasoning are presented in
Tables~\ref{table:result1} and \ref{table:result2}. It can be seen that HiSMatch
consistently outperforms all the baselines on all the six TKGs, which indicates
its effectiveness and superiority. Especially on ICEWS14 and ICEWS05-15,
HiSMatch achieves the most significant improvements of 5.6\% and 4.8\% in MRR,
respectively. In more detail, we have the following observations: (1) HiSMatch
outperforms all the KG reasoning models because it can capture both the time
information for each fact and sequential patterns in TKGs; (2) HiSMatch performs
much better than those interpolation models because they cannot learn
representations for unseen timestamps; (3) More importantly, HiSMatch gets
better results than all the extrapolation baselines, which proves the
superiority of modeling both two kinds of history, i.e., query-related history and
candidate-related history; (4) It can be seen that the baselines (e.g., TITer)
focusing on query-related history are usually strong on precision and get good
results on $Hits@1$ while the baselines focus on the candidate-related history
(e.g., RE-GCN) are more capable on recall and get good results on $Hits@10$. In a
word, by transforming the TKG reasoning task into a matching task, HiSMatch
utilizes both two kinds of history more comprehensively. Moreover, HiSMatch captures
more high-order associations via the background knowledge graph. Therefore, it
gets the best performances in all the metrics. 

By conducting experiments on six datasets with different time granularities, we
found that the time granularity partly determines what is vital to the TKG
reasoning task. Take the two most typical datasets for example, (1) GDELT has
the most fine-grained time granularity (15 minutes) and the results of all the
baselines are similarly poor, compared with those of the other datasets. There
are more timestamps in history when the time granularity gets more fine-grained,
which requires the model to capture history at more timestamps. Under the
matching framework, HiSMatch can capture longer historical information than
candidate-based models and more comprehensive history than query-based models.
Thus, it gets better results (2.3\% in MRR); (2) Contrary to GDELT, WIKI has the
largest time granularity (1 year). In this situation, the behavioral trends
implied in history at fewer timestamps are vital for the reasoning. Moreover,
there are more structural dependencies in each KG due to the large time
granularity. Thus, RE-GCN focuses on modeling the global structure at the latest
a few timestamps and gets strong performance on this data. Still, HiSMatch
outperforms it by modeling the two kinds of substructures and the background
knowledge.

\subsection{Ablation Study}
To further analyze how each part of HiSMatch contributes to the final results,
we report the MRR results of the HiSMatch variants on the validation sets on
three typical datasets, namely, ICEWS14, ICEWS18 and WIKI, in
Table~\ref{table:ablation}.

\textbf{Impact of the Query Structure Encoder.}
To demonstrate how the query structure encoder contributes to the final results
of HiSMatch, we remove the query structure encoder and use the representation of
the query entity from the candidate structure encoder as the representation of
the query. The results are denoted as \textit{-query} in
Table~\ref{table:ablation}. It can be seen that \textit{-query} performs
consistently worse than HiSMatch on all the datasets. It is because that
query-related historical structure can model the query more accurately by
modeling the repetitive facts focused on the query relation. 

\textbf{Impact of the Candidate Structure Encoder.}
The results denoted as \textit{-candidate} in Table~\ref{table:ablation}
demonstrate the performance of HiSMatch without modeling the candidate-related
history. More specifically, we directly add a fully connection layer after the
query structure encoder to get the scores of all entities
following~\cite{jin2020recurrent}. It can be observed that ignoring the
candidate-related historical structure has a great impact on the results.
Candidate-related history contains rich information that describes the
behavioral trends about all candidate entities, which is helpful to select the
correct answer. Especially on WIKI, the dataset with the largest time
granularity as mentioned in Section~\ref{exp_result}, entities have more
associations among each other at each timestamp and thus contain rich behaviors.

\textbf{Impact of the Background Knowledge Encoder.}
\textit{-background} in Table~\ref{table:ablation} denotes a variant of HiSMatch
that uses the learned representations of entities without the background
knowledge encoder. Note that, in the training phase, the randomly initialized
representations of entities will be learned and updated. In the test phase, the
model uses the learned representations of entities as the input. It can be
observed that the performance of \textit{-background} is worse than HiSMatch on
all the datasets, especially on the WIKI, which has a time granularity of one
year. There are more high order associations in the background knowledge graph.
Thus, the background knowledge is more important for WIKI than other datesets.

\textbf{Impact of the Time Semantic Component.}
To demonstrate how the time semantic component contributes to the final results,
we remove the time semantic component and only use the outputs of structure
semantic component as the inputs of the sequential pattern component. The
results are denoted as \textit{-time} in Table~\ref{table:ablation}. It can be
seen that the time semantic component is useful on all the datasets. It is
because the time semantic component describes the time numerical information so
that it can help HisMatch to distinguish different time intervals between the
history and the query. 



\begin{table}
  \scriptsize 
  \centering
  \setlength{\tabcolsep}{0.3em}
  \begin{tabular}{llllll}
  \toprule
  Model   & ICE14  & ICE18  &WIKI\\
  \midrule
  HiSMatch &\bf{47.89} &\bf{35.18}  &\bf{79.91}\\

  \textit{-query}      &44.86 (-3.0)    &33.16 (-2.0)    &77.36  (-2.6)          \\
  \textit{-cantidate}     &40.04 (-7.9)    &29.12 (-6.1)    &63.26 (-16.7)        \\
  \textit{-background} &44.26 (-3.6)    &32.87 (-2.3)    &73.54 (-6.4)         \\
  \textit{-time}       &45.01 (-2.9)    &33.25 (-1.9)    &78.60 (-1.3)   \\
  \bottomrule
  \end{tabular}
  \caption{MRR results (in percentage) by different variants of HiSMatch on three datasets.}
  \label{table:ablation}
  \end{table}

  \begin{table}[htbp]
    \scriptsize 
    \centering
    \begin{tabular}{llllll}
    \toprule
    Model   & ICE14  & ICE18  &WIKI\\
    \midrule
    HiSMatch (CompGCN)\!\!\!    &\bf{47.89} &\bf{35.18} &\textbf{79.91}  \\ 
    HiSMatch (CompGCN-mult)\!\!\!  &46.12  &34.45&73.45 \\ 
    HiSMatch (RGCN)\!\!\!    &47.03  &35.08  &74.83              \\ 
    HiSMatch (KBAT)\!\!\!    &47.53  &34.78 &77.12 \\ 
    \bottomrule
    \end{tabular}
    \caption{Performance (in percentage) of HisMatch with different kinds of GCNs.}
    \label{hismatch_gcn}
    \end{table}

  \subsection{Comparative Study on Different GCNs}
  \label{sec: gcn}
  To further study the impact of different kinds of GCNs in the candidate
  structure encoder and the background knowledge encoder, we replace CompGCN in
  these two encoders with CompGCN-mult~\cite{vashishth2019composition},
  RGCN~\cite{schlichtkrull2018modeling} and KBAT~\cite{nathani2019learning}. The
  MRR results on the validation sets of ICE14, ICE18, and WIKI are reported in
  Table~\ref{hismatch_gcn}. It can be seen that HiSMatch (CompGCN) gets the best
  performance. For ICE14 and ICE18, the two datasets with the time granularities
  of one day, the structure dependencies are relatively simple. Thus, different
  GCNs get similar performances. While for WIKI, the dateset with the time
  granularities of one year, there are more structural dependencies in the
  candidate-related history and the background knowledge graph. Therefore, the
  performance gap caused by the capabilities of GCNs becomes more significant.

\begin{figure}[tbp] 
  \centering
  \includegraphics[width=3in]{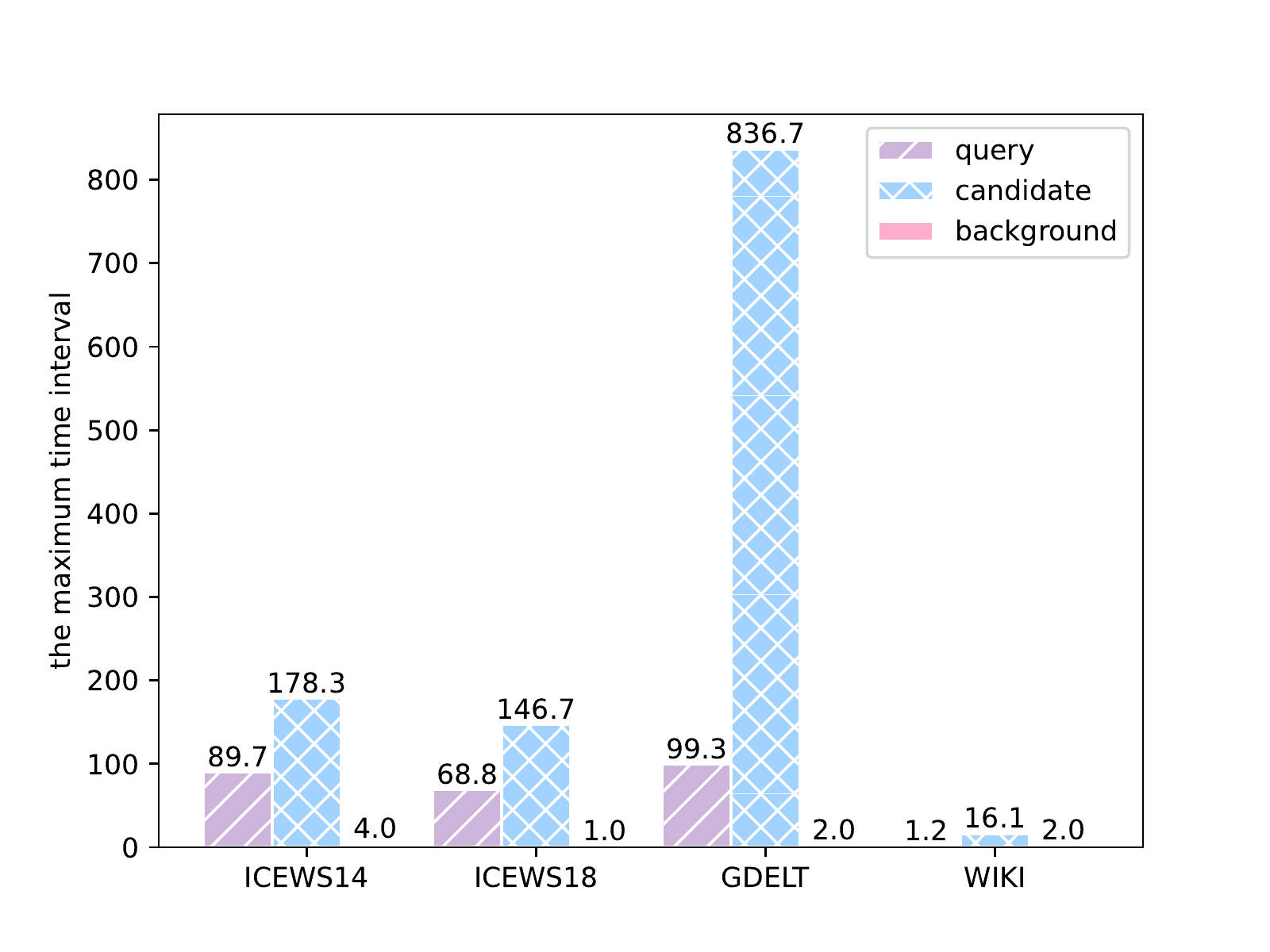}
  \caption{Statistic of maximum time intervals in history on four datasets.}
  \label{fig:history-len}
  \vspace{-6mm}
  \end{figure}
  
\subsection{Study on Maximum Time Interval}
\label{sec: detail analysis}
To explore the maximum time intervals between the query time and the history
that HiSMatch models, we conduct statistics on the maximum time interval of
historical facts in the query-related histoical structure ($\Delta t= t_q-t_1$)
and candidate-related historical structure ($\Delta t'=t_q-t'_1$) under the
optimal parameters (We report the average maximum time interval on the
validation sets). We also report the time interval of background knowledge
graphs ($k$) for comparison. As shown in Figure~\ref{fig:history-len}, $\Delta
t$ and $\Delta t'$ are both much larger than $k$. The results demonstrate that
the two historical structures can modeling the long-term behaviors of the query
and candidates. The background knowledge graph focuses on model high-order
associations among all the facts at the latest a few timestamps, which models
the global structure dependencies in a much shorter time interval. It can be
seen that $\Delta t$ on GDELT is more than 800 and the value is only around 16
on WIKI, which verifies the discussion in Section~\ref{exp_result}.

\section{Conclusion}
In this paper, we considered both two kinds of history, namely, the
query-related history and the candidate-related history in TKG reasoning and
transformed the task into a matching problem between them for the first time. We
further proposed the HiSMatch model, which applies two structure encoders to
calculate the representations of historical structures of the queries and
candidates, respectively. Each encoder contains a structure semantic component
to model the concurrent structure among entities, a time semantic component to
model the time numerical information of facts, and a sequence pattern component
to capture the temporal orders. Besides, HiSMatch integrates the background
knowledge into the representations of entities. Experimental results on six
benchmark datasets demonstrate the superiority of HiSMatch.

\section{Limitations}

The limitations of this work can be concluded into two points: (1) HiSMatch uses
a heuristic history finding strategy to get two kinds of history, which may lose
some critical facts. Although it uses the background knowledge encoder to
consider more historical facts, a learnable history finding strategy is more
helpful. (2) HiSMatch is an initial attempt to apply the matching framework to
solve the TKG reasoning task using two separate encoders for each kind of
history, which fails to model the interactions between the two kinds of history
explicitly. Designing a cross-encoder to match history more comprehensively is a
good direction for future studies. 

\section*{Acknowledgments}
The work is supported by the National Natural Science Foundation of China under
grants U1911401, 62002341 and 61772501, the GFKJ Innovation Program, Beijing
Academy of Artificial Intelligence under grant BAAI2019ZD0306, and the
Lenovo-CAS Joint Lab Youth Scientist Project.

\bibliography{anthology,custom}

\begin{thebibliography}{36}
\expandafter\ifx\csname natexlab\endcsname\relax\def\natexlab#1{#1}\fi

\bibitem[{Bordes et~al.(2013)Bordes, Usunier, Garcia-Duran, Weston, and
  Yakhnenko}]{bordes2013translating}
Antoine Bordes, Nicolas Usunier, Alberto Garcia-Duran, Jason Weston, and Oksana
  Yakhnenko. 2013.
\newblock Translating embeddings for modeling multi-relational data.
\newblock \emph{Advances in neural information processing systems}, 26.

\bibitem[{Dasgupta et~al.(2018)Dasgupta, Ray, and Talukdar}]{dasgupta2018hyte}
Shib~Sankar Dasgupta, Swayambhu~Nath Ray, and Partha Talukdar. 2018.
\newblock Hyte: Hyperplane-based temporally aware knowledge graph embedding.
\newblock In \emph{Proceedings of the 2018 conference on empirical methods in
  natural language processing}, pages 2001--2011.

\bibitem[{Deng et~al.(2020)Deng, Rangwala, and Ning}]{deng2020dynamic}
Songgaojun Deng, Huzefa Rangwala, and Yue Ning. 2020.
\newblock Dynamic knowledge graph based multi-event forecasting.
\newblock In \emph{Proceedings of the 26th ACM SIGKDD International Conference
  on Knowledge Discovery \& Data Mining}, pages 1585--1595.

\bibitem[{Dettmers et~al.(2018)Dettmers, Minervini, Stenetorp, and
  Riedel}]{dettmers2018convolutional}
Tim Dettmers, Pasquale Minervini, Pontus Stenetorp, and Sebastian Riedel. 2018.
\newblock Convolutional 2d knowledge graph embeddings.
\newblock In \emph{Proceedings of the AAAI Conference on Artificial
  Intelligence}, volume~32.

\bibitem[{Garcia-Duran et~al.(2018)Garcia-Duran, Duman{\v{c}}i{\'c}, and
  Niepert}]{garcia2018learning}
Alberto Garcia-Duran, Sebastijan Duman{\v{c}}i{\'c}, and Mathias Niepert. 2018.
\newblock Learning sequence encoders for temporal knowledge graph completion.
\newblock In \emph{Proceedings of the 2018 Conference on Empirical Methods in
  Natural Language Processing}, pages 4816--4821.

\bibitem[{Goel et~al.(2020)Goel, Kazemi, Brubaker, and
  Poupart}]{goel2020diachronic}
Rishab Goel, Seyed~Mehran Kazemi, Marcus Brubaker, and Pascal Poupart. 2020.
\newblock Diachronic embedding for temporal knowledge graph completion.
\newblock In \emph{Proceedings of the AAAI Conference on Artificial
  Intelligence}, volume~34, pages 3988--3995.

\bibitem[{Han et~al.(2020{\natexlab{a}})Han, Chen, Ma, and
  Tresp}]{han2020dyernie}
Zhen Han, Peng Chen, Yunpu Ma, and Volker Tresp. 2020{\natexlab{a}}.
\newblock Dyernie: Dynamic evolution of riemannian manifold embeddings for
  temporal knowledge graph completion.
\newblock In \emph{Proceedings of the 2020 Conference on Empirical Methods in
  Natural Language Processing (EMNLP)}, pages 7301--7316.

\bibitem[{Han et~al.(2020{\natexlab{b}})Han, Chen, Ma, and
  Tresp}]{han2020explainable}
Zhen Han, Peng Chen, Yunpu Ma, and Volker Tresp. 2020{\natexlab{b}}.
\newblock Explainable subgraph reasoning for forecasting on temporal knowledge
  graphs.
\newblock In \emph{International Conference on Learning Representations}.

\bibitem[{Han et~al.(2021{\natexlab{a}})Han, Ding, Ma, Gu, and
  Tresp}]{han2021learning}
Zhen Han, Zifeng Ding, Yunpu Ma, Yujia Gu, and Volker Tresp.
  2021{\natexlab{a}}.
\newblock Learning neural ordinary equations for forecasting future links on
  temporal knowledge graphs.
\newblock In \emph{Proceedings of the 2021 Conference on Empirical Methods in
  Natural Language Processing}, pages 8352--8364.

\bibitem[{Han et~al.(2021{\natexlab{b}})Han, Ding, Ma, Gu, and
  Tresp}]{han-etal-2021-learning-neural}
Zhen Han, Zifeng Ding, Yunpu Ma, Yujia Gu, and Volker Tresp.
  2021{\natexlab{b}}.
\newblock \href {https://doi.org/10.18653/v1/2021.emnlp-main.658} {Learning
  neural ordinary equations for forecasting future links on temporal knowledge
  graphs}.
\newblock In \emph{Proceedings of the 2021 Conference on Empirical Methods in
  Natural Language Processing}, pages 8352--8364, Online and Punta Cana,
  Dominican Republic. Association for Computational Linguistics.

\bibitem[{Han et~al.(2020{\natexlab{c}})Han, Ma, Wang, G{\"u}nnemann, and
  Tresp}]{han2020graph}
Zhen Han, Yunpu Ma, Yuyi Wang, Stephan G{\"u}nnemann, and Volker Tresp.
  2020{\natexlab{c}}.
\newblock Graph hawkes neural network for forecasting on temporal knowledge
  graphs.
\newblock \emph{arXiv preprint arXiv:2003.13432}.

\bibitem[{He et~al.(2017)He, Balakrishnan, Eric, and Liang}]{he2017learning}
He~He, Anusha Balakrishnan, Mihail Eric, and Percy Liang. 2017.
\newblock Learning symmetric collaborative dialogue agents with dynamic
  knowledge graph embeddings.
\newblock \emph{arXiv preprint arXiv:1704.07130}.

\bibitem[{Jiang et~al.(2016)Jiang, Liu, Ge, Sha, Li, Chang, and
  Sui}]{jiang2016encoding}
Tingsong Jiang, Tianyu Liu, Tao Ge, Lei Sha, Sujian Li, Baobao Chang, and
  Zhifang Sui. 2016.
\newblock Encoding temporal information for time-aware link prediction.
\newblock In \emph{Proceedings of the 2016 Conference on Empirical Methods in
  Natural Language Processing}, pages 2350--2354.

\bibitem[{Jin et~al.(2019)Jin, Qu, Jin, and Ren}]{jin2019recurrent}
Woojeong Jin, Meng Qu, Xisen Jin, and Xiang Ren. 2019.
\newblock Recurrent event network: Autoregressive structure inference over
  temporal knowledge graphs.
\newblock \emph{arXiv preprint arXiv:1904.05530}.

\bibitem[{Jin et~al.(2020)Jin, Qu, Jin, and Ren}]{jin2020recurrent}
Woojeong Jin, Meng Qu, Xisen Jin, and Xiang Ren. 2020.
\newblock Recurrent event network: Autoregressive structure inferenceover
  temporal knowledge graphs.
\newblock In \emph{Proceedings of the 2020 Conference on Empirical Methods in
  Natural Language Processing (EMNLP)}, pages 6669--6683.

\bibitem[{Kingma and Ba(2014)}]{kingma2014adam}
Diederik~P Kingma and Jimmy Ba. 2014.
\newblock Adam: A method for stochastic optimization.
\newblock \emph{arXiv preprint arXiv:1412.6980}.

\bibitem[{Lacroix et~al.(2020)Lacroix, Obozinski, and
  Usunier}]{lacroix2020tensor}
Timoth{\'e}e Lacroix, Guillaume Obozinski, and Nicolas Usunier. 2020.
\newblock Tensor decompositions for temporal knowledge base completion.
\newblock \emph{arXiv preprint arXiv:2004.04926}.

\bibitem[{Lan and Jiang(2020)}]{lan2020query}
Yunshi Lan and Jing Jiang. 2020.
\newblock Query graph generation for answering multi-hop complex questions from
  knowledge bases.
\newblock Association for Computational Linguistics.

\bibitem[{Leblay and Chekol(2018)}]{leblay2018deriving}
Julien Leblay and Melisachew~Wudage Chekol. 2018.
\newblock Deriving validity time in knowledge graph.
\newblock In \emph{Companion Proceedings of the The Web Conference 2018}, pages
  1771--1776.

\bibitem[{Leetaru and Schrodt(2013)}]{leetaru2013gdelt}
Kalev Leetaru and Philip~A Schrodt. 2013.
\newblock Gdelt: Global data on events, location, and tone, 1979--2012.
\newblock In \emph{ISA annual convention}, volume~2, pages 1--49. Citeseer.

\bibitem[{Li et~al.(2022)Li, Guan, Jin, Peng, Lyu, Zhu, Bai, Li, Guo, and
  Cheng}]{li-etal-2022-complex}
Zixuan Li, Saiping Guan, Xiaolong Jin, Weihua Peng, Yajuan Lyu, Yong Zhu, Long
  Bai, Wei Li, Jiafeng Guo, and Xueqi Cheng. 2022.
\newblock \href {https://doi.org/10.18653/v1/2022.acl-short.32} {Complex
  evolutional pattern learning for temporal knowledge graph reasoning}.
\newblock In \emph{Proceedings of the 60th Annual Meeting of the Association
  for Computational Linguistics (Volume 2: Short Papers)}, pages 290--296,
  Dublin, Ireland. Association for Computational Linguistics.

\bibitem[{Li et~al.(2021{\natexlab{a}})Li, Jin, Guan, Li, Guo, Wang, and
  Cheng}]{li2021search}
Zixuan Li, Xiaolong Jin, Saiping Guan, Wei Li, Jiafeng Guo, Yuanzhuo Wang, and
  Xueqi Cheng. 2021{\natexlab{a}}.
\newblock Search from history and reason for future: Two-stage reasoning on
  temporal knowledge graphs.
\newblock \emph{arXiv preprint arXiv:2106.00327}.

\bibitem[{Li et~al.(2021{\natexlab{b}})Li, Jin, Li, Guan, Guo, Shen, Wang, and
  Cheng}]{li2021temporal}
Zixuan Li, Xiaolong Jin, Wei Li, Saiping Guan, Jiafeng Guo, Huawei Shen,
  Yuanzhuo Wang, and Xueqi Cheng. 2021{\natexlab{b}}.
\newblock Temporal knowledge graph reasoning based on evolutional
  representation learning.
\newblock In \emph{Proceedings of the 44th International ACM SIGIR Conference
  on Research and Development in Information Retrieval}, pages 408--417.

\bibitem[{Liao et~al.(2021)Liao, Liang, Meng, and Zhang}]{liao2021learning}
Siyuan Liao, Shangsong Liang, Zaiqiao Meng, and Qiang Zhang. 2021.
\newblock Learning dynamic embeddings for temporal knowledge graphs.
\newblock In \emph{Proceedings of the 14th ACM International Conference on Web
  Search and Data Mining}, pages 535--543.

\bibitem[{Nathani et~al.(2019)Nathani, Chauhan, Sharma, and
  Kaul}]{nathani2019learning}
Deepak Nathani, Jatin Chauhan, Charu Sharma, and Manohar Kaul. 2019.
\newblock Learning attention-based embeddings for relation prediction in
  knowledge graphs.
\newblock In \emph{Proceedings of the 57th Annual Meeting of the Association
  for Computational Linguistics}, pages 4710--4723.

\bibitem[{Schlichtkrull et~al.(2018)Schlichtkrull, Kipf, Bloem, Berg, Titov,
  and Welling}]{schlichtkrull2018modeling}
Michael Schlichtkrull, Thomas~N Kipf, Peter Bloem, Rianne van~den Berg, Ivan
  Titov, and Max Welling. 2018.
\newblock Modeling relational data with graph convolutional networks.
\newblock In \emph{European semantic web conference}, pages 593--607. Springer.

\bibitem[{Shang et~al.(2019)Shang, Tang, Huang, Bi, He, and
  Zhou}]{shang2019end}
Chao Shang, Yun Tang, Jing Huang, Jinbo Bi, Xiaodong He, and Bowen Zhou. 2019.
\newblock End-to-end structure-aware convolutional networks for knowledge base
  completion.
\newblock In \emph{Proceedings of the AAAI Conference on Artificial
  Intelligence}, volume~33, pages 3060--3067.

\bibitem[{Sun et~al.(2021)Sun, Zhong, Ma, Han, and He}]{sun2021timetraveler}
Haohai Sun, Jialun Zhong, Yunpu Ma, Zhen Han, and Kun He. 2021.
\newblock Timetraveler: Reinforcement learning for temporal knowledge graph
  forecasting.
\newblock In \emph{Proceedings of the 2021 Conference on Empirical Methods in
  Natural Language Processing}, pages 8306--8319.

\bibitem[{Sun et~al.(2018)Sun, Deng, Nie, and Tang}]{sun2018rotate}
Zhiqing Sun, Zhi-Hong Deng, Jian-Yun Nie, and Jian Tang. 2018.
\newblock Rotate: Knowledge graph embedding by relational rotation in complex
  space.
\newblock In \emph{International Conference on Learning Representations}.

\bibitem[{Trouillon et~al.(2016)Trouillon, Welbl, Riedel, Gaussier, and
  Bouchard}]{trouillon2016complex}
Th{\'e}o Trouillon, Johannes Welbl, Sebastian Riedel, {\'E}ric Gaussier, and
  Guillaume Bouchard. 2016.
\newblock Complex embeddings for simple link prediction.
\newblock In \emph{International conference on machine learning}, pages
  2071--2080. PMLR.

\bibitem[{Vashishth et~al.(2019)Vashishth, Sanyal, Nitin, and
  Talukdar}]{vashishth2019composition}
Shikhar Vashishth, Soumya Sanyal, Vikram Nitin, and Partha Talukdar. 2019.
\newblock Composition-based multi-relational graph convolutional networks.
\newblock In \emph{International Conference on Learning Representations}.

\bibitem[{Wang et~al.(2019)Wang, He, Cao, Liu, and Chua}]{wang2019kgat}
Xiang Wang, Xiangnan He, Yixin Cao, Meng Liu, and Tat-Seng Chua. 2019.
\newblock Kgat: Knowledge graph attention network for recommendation.
\newblock In \emph{Proceedings of the 25th ACM SIGKDD International Conference
  on Knowledge Discovery \& Data Mining}, pages 950--958.

\bibitem[{Wu et~al.(2020)Wu, Cao, Cheung, and Hamilton}]{wu2020temp}
Jiapeng Wu, Meng Cao, Jackie Chi~Kit Cheung, and William~L Hamilton. 2020.
\newblock Temp: Temporal message passing for temporal knowledge graph
  completion.
\newblock In \emph{Proceedings of the 2020 Conference on Empirical Methods in
  Natural Language Processing (EMNLP)}, pages 5730--5746.

\bibitem[{Xu et~al.(2021)Xu, Sun, Zhang, Yi, Miao, Yang, Meng, Hu, Wang, Min
  et~al.}]{xu2021time}
Yonghui Xu, Shengjie Sun, Huiguo Zhang, Chang’an Yi, Yuan Miao, Dong Yang,
  Xiaonan Meng, Yi~Hu, Ke~Wang, Huaqing Min, et~al. 2021.
\newblock Time-aware graph embedding: A temporal smoothness and task-oriented
  approach.
\newblock \emph{ACM Transactions on Knowledge Discovery from Data (TKDD)},
  16(3):1--23.

\bibitem[{Yang et~al.(2015)Yang, Yih, He, Gao, and Deng}]{yang2015embedding}
Bishan Yang, Scott Wen-tau Yih, Xiaodong He, Jianfeng Gao, and Li~Deng. 2015.
\newblock Embedding entities and relations for learning and inference in
  knowledge bases.
\newblock In \emph{Proceedings of the International Conference on Learning
  Representations (ICLR) 2015}.

\bibitem[{Zhu et~al.(2021)Zhu, Chen, Fan, Cheng, and Zhang}]{zhu2021learning}
Cunchao Zhu, Muhao Chen, Changjun Fan, Guangquan Cheng, and Yan Zhang. 2021.
\newblock Learning from history: Modeling temporal knowledge graphs with
  sequential copy-generation networks.
\newblock In \emph{Proceedings of the AAAI Conference on Artificial
  Intelligence}, volume~35, pages 4732--4740.

\end{thebibliography}
\bibliographystyle{acl_natbib}

\newpage
\appendix

\end{document}